\documentclass[11pt,a4paper]{article}
\usepackage[hyperref]{emnlp2017}
\usepackage{times}
\usepackage{latexsym}

\usepackage{url}
\usepackage{alltt}

\usepackage{listings}
\usepackage{amsmath}
\usepackage{verbatim}
\usepackage{tikz-qtree}
\usepackage{tikz}

\usepackage{filecontents}
\usepackage{pgfplots, pgfplotstable}
\usepgfplotslibrary{statistics}

\emnlpfinalcopy 
\def\emnlppaperid{820} 

\newcommand{\ignore}[1]{}

\newsavebox{\one}
\newsavebox{\two}
\newsavebox{\three}
\newsavebox{\four}
\newsavebox{\five}

\ifx\release\undefined
    
\fi

\title{AMR Parsing using Stack-LSTMs}

\author{Miguel Ballesteros~~~~~~~~~~Yaser Al-Onaizan\\
IBM T.J Watson Research Center, \\ 1101 Kitchawan Road, Route 134
Yorktown Heights, NY 10598. U.S \\
{ \sf miguel.ballesteros@ibm.com, onaizan@us.ibm.com } }
\date{}

\begin{document}
\maketitle
\begin{abstract}
We present a transition-based AMR parser that directly generates AMR parses from plain text. 
  We use Stack-LSTMs to represent 
  our parser state and make decisions greedily. In our experiments, we show that our parser achieves 
  very competitive scores on English using only AMR training data. Adding additional 
information, such as POS tags and dependency trees, improves the results further.
\end{abstract}

\section{Introduction}

Transition-based algorithms for natural language parsing \cite{yamada03,nivre03iwpt,arcstd,nivre08cl}
are formulated as a series of decisions that read words from a buffer and incrementally combine 
them to form syntactic structures in a stack. Apart from dependency parsing, these models, also known as 
shift-reduce algorithms, have been successfully applied to tasks like phrase-structure parsing 
\cite{ZhangC11,cdyer2016},  named entity recognition \cite{lample2016},  CCG parsing \cite{misra-artzi:2016:EMNLP2016}
joint syntactic and 
semantic parsing \cite{henderson2013,SwayamdiptaBDS16} and 
even abstract-meaning representation parsing \cite{wangxuepradhan,WangXP15,DamonteCS16}.

AMR parsing requires solving several natural language processing tasks; mainly named entity 
recognition, word sense disambiguation and 
joint syntactic and semantic role  labeling.\footnote{Check \cite{banarescu} for a complete description of 
AMR graphs.} 
Given the difficulty of building an end-to-end system, most prior work  is based on 
pipelines  or heavily dependent on precalculated features \cite[inter-alia]{jamr,zhouemnlp,stanfordamr,wangxuepradhan}.

Inspired by \newcite{wangxuepradhan,WangXP15,goodman-vlachos-naradowsky:2016:P16-1,DamonteCS16} and \newcite{lstmacl15}, we 
present a shift-reduce algorithm that produces AMR graphs directly from plain text. 
\newcite{wangxuepradhan,WangXP15,zhouemnlp,goodman-vlachos-naradowsky:2016:P16-1} presented 
transition-based tree-to-graph transducers that traverse a dependency tree and 
transforms it to an AMR graph. \newcite{DamonteCS16}'s input is a sentence and it is therefore more similar 
(with a different parsing algorithm) to our approach, but their parser relies on 
external tools, such as dependency parsing, semantic role labeling or named entity recognition.

The input of our parser is plain text sentences and, through rich word representations,  it predicts 
all actions (in a single algorithm) needed to 
generate an AMR graph representation for an input sentence; it handles the 
detection and annotation of named entities, word sense disambiguation and it 
makes connections between the nodes detected towards building a predicate 
argument structure. Even though the system that runs with just words is very competitive, 
we further improve the results incorporating POS tags and dependency trees into our model.

Stack-LSTMs\footnote{We use 
the dynamic framework of \newcite{dynet} to implement our parser.} have proven to be useful in 
tasks related to syntactic and semantic parsing 
\cite{lstmacl15,cdyer2016,SwayamdiptaBDS16} and named entity 
recognition \cite{lample2016}. In this paper, we demonstrate that they can be effectively used for AMR 
parsing as well.

\section{Parsing Algorithm}
\label{algorithm}

Our parsing algorithm makes use of a \textsc{stack} (that stores AMR nodes and/or words) and 
a \textsc{buffer} that contains the words that have yet to be processed.  The parsing algorithm 
is inspired from the semantic actions presented by \newcite{henderson2013}, the 
transition-based NER algorithm by \newcite{lample2016} and the arc-standard algorithm  \cite{arcstd}. 
As in \cite{BallesterosN13} the buffer starts with the root symbol at the end of the sequence. 
Figure \ref{parsingexample} shows  a running example. 
The transition inventory is the following: 
\begin{itemize}

  \item \textsc{shift}: pops the front of the \textsc{buffer} and push it to the \textsc{stack}.
  
  \item \textsc{confirm}: calls a subroutine that predicts the AMR node corresponding to the top of 
  the \textsc{stack}. It then pops the word from the \textsc{stack} and pushes the 
  AMR node to the \textsc{stack}. An example is the prediction of a propbank sense: 
  From \emph{occured} \textbf{to} \emph{occur-01}.
  
    \item \textsc{reduce}: pops the top of the \textsc{stack}. It  
  occurs when the word/node at the top of the  stack is complete 
  (no more actions can be applied to it). Note that it can also be applied to words that do not 
  appear in the final output graph, and thus they are directly discarded.
  
  \item  \textsc{merge}: pops the two nodes at the top of the \textsc{stack} and then it merges them, 
  it then pushes the resulting node to the top of \textsc{stack}. Note that this can be applied recursively. 
  This  action serves to get multiword named entities (e.g. \emph{New York 
  City}).
  
  \item \textsc{Entity}(label): labels the node at the top of the \textsc{stack} 
  with an entity label. This action serves to label named entities, such as  \emph{New York City} or 
  \emph{Madrid} and it is normally run after \textsc{Merge} when it is a 
  multi-word named entity, or after \textsc{SHIFT} if it is a single-word named 
  entity.

  \item \textsc{dependent}(label,node): creates a new node in the AMR graph that is dependent on the 
  node at the top of the \textsc{stack}. An example is the 
  introduction of a negative \emph{polarity} to a given node: 
  From \emph{illegal} \textbf{to} (\emph{legal, polarity -}).

  \item \textsc{LA}(label) and  \textsc{RA}(label): create a left/right arc with the top two 
  nodes at the top of the \textsc{stack}. They keep both the head and the dependent in the stack to allow 
  reentrancies (multiple incoming edges). The head is now a composition of the head and the dependent. 
  They are enriched with the AMR label.
    
   \item \textsc{SWAP}:  pops the two top items at the top of the \textsc{stack}, pushes the second node to 
   the front of the \textsc{buffer}, and pushes the first one back into the
    \textsc{stack}.  This action allows non-projective arcs as in \cite{nivre2009non} 
    but it also helps to introduce reentrancies.
    At oracle time, \textsc{SWAP} is produced when the word at the top of the stack is blocking actions that 
    may happen between the second element at the top of the stack and any of the words in the buffer.    
\end{itemize}

Figure \ref{fig:parser} shows the parser actions and the effect on the parser 
state (contents of the stack, buffer) and how the graph is changed after 
applying the actions.

\begin{figure*}
\centering
\begin{tabular}{cc|l|cc|c}
\textbf{Stack}$_t$ & \textbf{Buffer}$_t$ & \textbf{Action} & \textbf{Stack}$_{t+1}$ & \textbf{Buffer}$_{t+1}$ & \textbf{Graph} \\
\hline
$S$ & $u,B$ & \textsc{shift} & $u,S$ & $B$ & --- \\ 
$u, S$ & $B$  &$\textsc{confirm}$ & $n,S$ & $B$ & ---   \\
$u, S$ & $B$  &$\textsc{reduce}$ & $S$ & $B$ & --- \\
$u, v,S$ & $B$  &$\textsc{merge}$ & $(u, v),S$ & $B$ &  ---   \\
$u, S$ & $B$  &$\textsc{entity}(l)$ & $(u:l),S$ & $B$ & --- \\
$u, S$ & $B$  &$\textsc{dependent}(r, d)$ & $u,S$ & $B$ & $u \stackrel{\scriptsize{r}}{\rightarrow} d $  \\
$u,v,S$ & $B$  &$\textsc{ra}(r)$ & $u,v,S$ & $B$ & $u \stackrel{\scriptsize{r}}{\rightarrow} v$   \\
$u,v, S$ & $B$ & $\textsc{la}(r)$ & $u,v, S$ & $B$ & $u \stackrel{\scriptsize{r}}{\leftarrow} v$  \\
$u,v,S$ & $B$ & \textsc{swap} & $u,S$ & $v,B$ & ---   
\end{tabular}
\caption{\label{fig:parser}Parser transitions indicating the action applied to the stack and buffer and 
the resulting state. }
\end{figure*}
 

 \begin{figure}[!ht]

  \begin{center}
    \centering
    \begin{tiny}
      
   \setlength{\tabcolsep}{0.85pt}     
\renewcommand{\arraystretch}{1}
      \begin{tabular}{l|l|l}\textsc{Action}&\textsc{Stack}&\textsc{Buffer}\\
      \hline
               INIT &&It, should, be, vigorously, advocated, R\\
               SHIFT &it&should, be, vigorously, advocated, R\\
               CONFIRM &it&should, be, vigorously, advocated, R \\
               SHIFT &should, it&be, vigorously, advocated, , R \\
               CONFIRM &recommend-01, it&be, vigorously, advocated, R \\
               SWAP&recommend-01&it, be, vigorously, advocated, R\\
               SHIFT &it, recommend-01&be, vigorously, advocated, R \\
               SHIFT&be, it, recommend-01&vigorously, advocated, R \\
               REDUCE &it, recommend-01&vigorously, advocated, R \\
               SHIFT &vigorously, it, recommend-01&advocated, R \\
               CONFIRM &vigorous, it, recommend-01&advocated, R \\
               SWAP &vigorous, recommend-01&it, advocated, R\\
               SWAP&vigorous&recommend-01, it, advocated, R \\
               SHIFT &recommend-01, vigorous&it, advocated, R \\
               SHIFT &it, recommend-01, vigorous&advocated , R\\
               SHIFT  & it, recommend-01, vigorous&advocated, R\\
               SHIFT &advocated, it, recommend-01, vigorous& R\\
               CONFIRM  &advocate-01, it, recommend-01, vigorous& R\\
               LA(ARG1)  &advocate-01, it, recommend-01, vigorous& R\\
               SWAP &advocate-01,  recommend-01, vigorous&it R\\
               SHIFT &it, advocate-01,  recommend-01, vigorous& R\\
               REDUCE &advocate-01,  recommend-01, vigorous& R\\
               RA(ARG1) & advocate-01, recommend-01, vigorous& R\\
               SWAP & advocate-01, vigorous & recommend-01, R\\
               SHIFT & recommend01, advocate-01, vigorous &  R\\
               SHIFT & R, recommend01, advocate-01, vigorous &  \\
               LA(root) & R, recommend01, advocate-01, vigorous &  \\
               REDUCE & recommend01, advocate-01, vigorous &  \\
               REDUCE & advocate-01, vigorous &\\
               LA(manner) & advocate-01, vigorous &  \\
               REDUCE & vigorous & \\
               REDUCE  &  & \\
                \vspace{-0.3cm}
      \end{tabular}
   \end{tiny}
    \vspace{1em}

    \begin{center}
      \begin{scriptsize}
      \begin{verbatim}
            (r / recommend-01 
                :ARG1 (a / advocate-01 
                      :ARG1 (i / it) 
                      :manner (v / vigorous)))   
                       
\end{verbatim}
 \vspace{-0.6cm}
\end{scriptsize}
\end{center}
    \caption{Transition sequence for the sentence \emph{It should be vigorously advocated}. 
    R represents the root symbol  }
    \label{parsingexample}
  \end{center}
\end{figure}
  
We implemented an oracle that produces the sequence of actions that leads to the 
  gold (or close to gold) AMR graph. In order to map words in the sentences to nodes in the AMR graph we 
  need to align them. We use the JAMR aligner provided by 
  \newcite{jamr}.\footnote{We used the latest version of the aligner \cite{flanigan2016cmu}}
  It is important to mention that even though the 
  aligner is quite accurate, it is not perfect, producing a F1 score of around 0.90. This means that most 
  sentences have at least one alignment error which implies that our oracle is 
  not capable of perfectly reproducing all AMR graphs. This has a direct impact on the 
 accuracy of the parser described in the next section since it is trained on 
sequences of actions that are not perfect.
The oracle achieves 0.895 F1 Smatch score \cite{cai2013smatch} when it is run on the development
 set of the LDC2014T12.
   
 The algorithm allows a set of different constraints that varies from the basic 
 ones (not allowing impossible actions such as \textsc{SHIFT} when the buffer is empty or not generating
 arcs when the words have not yet been \textsc{confirm}ed and thus transformed to nodes) 
 to more complicated ones based on the propbank candidates and number of arguments. We choose 
 to constrain the parser to the basic ones and let it learn the more complicated ones.  
  
 \section{Parsing Model}

In this section, we revisit Stack-LSTMs, our parsing model and our word representations.

\subsection{Stack-LSTMs}


The \textbf{stack LSTM} is an augmented LSTM \cite{hochreiter:1997,graves:2013} that allows adding 
new inputs in the same way as LSTMs
but it also provides a \textsc{pop} operation that moves a pointer to the previous element.  
The output vector of the LSTM will consider the stack pointer instead of the rightmost position
 of the sequence.\footnote{We refer interested readers to \cite{lstmacl15} for further 
 details.}
 
 \subsection{Representing the State and Making Parsing Decisions}
 \label{staterep}
 
The state of the algorithm presented in Section \ref{algorithm} is represented by the contents of the 
\textsc{stack}, \textsc{buffer}  and a list with the history of actions (which are encoded as Stack-LSTMs).\footnote{
Word representations, input and hidden representations have 100 dimensions, action and label representations 
are of size 20.}
All of this forms the vector  $\mathbf{s}_t$ that represents the state which s calculated as follows:

\vspace{-0.25cm}

\begin{align*}
\mathbf{s}_t = \max \left\{\mathbf{0}, \mathbf{W}[\mathbf{st}_t; \mathbf{b}_t; \mathbf{a}_t] + \mathbf{d}\right\},
\end{align*}

\noindent where $\mathbf{W}$ is a learned parameter matrix, $\mathbf{d}$ is a bias term and $\mathbf{st}_t$,
$\mathbf{b}_t$,$\mathbf{a}_t$ represent the output vector of the Stack-LSTMs at 
time $t$.

\paragraph{Predicting the Actions:}
Our model then uses the vector $\mathbf{s}_t$ for each timestep $t$ to compute the 
probability of the next action as:

\vspace{-0.15cm}
\begin{align}
p(z_t \mid \mathbf{s}_t) = \frac{\exp \left( \mathbf{g}_{z_t}^{\top}
\mathbf{s}_t + q_{z_t} \right)}{\sum_{z' \in \mathcal{A}} \exp \left(
\mathbf{g}_{z'}^{\top} \mathbf{s}_t + q_{z'} \right)}
\label{eq:local-objective}
\end{align}


\noindent where $\mathbf{g}_z$ is a column vector representing the (output) embedding of the action $z$, 
and $q_z$ is a bias term for action $z$. The set $\mathcal{A}$ represents the actions 
listed in Section \ref{algorithm}. Note that due to parsing constraints the set of possible actions 
may vary. The total number of actions (in the LDC2014T12 dataset) is 478; note 
that they include all possible labels (in the case of \textsc{LA} and \textsc{RA} ) and the different dependent 
nodes for the \textsc{Dependent} action

\paragraph{Predicting the Nodes:}
When the model selects the action \textsc{confirm}, the model needs to 
decide the AMR node\footnote{When the word at the top of stack is an out of vocabulary word, 
the system directly outputs the word itself as AMR node.} that corresponds to the word at the top 
of the \textsc{stack}, by using  $\mathbf{s}_t$, as follows:
\vspace{-0.15cm}
\begin{align}
p(e_t \mid \mathbf{s}_t) = \frac{\exp \left( \mathbf{g}_{e_t}^{\top}
\mathbf{s}_t + q_{e_t} \right)}{\sum_{e' \in \mathcal{N}} \exp \left(
\mathbf{g}_{e'}^{\top} \mathbf{s}_t + q_{e'} \right)}
\label{eq:local-objective}
\end{align}

\noindent where $\mathcal{N}$ is 
the set of possible candidate nodes for the word at the top of the \textsc{stack}.  $\mathbf{g}_e$ is a column 
vector representing the (output) embedding of the node $e$, and $q_e$ is a bias term for the node 
$e$.  It is important to mention that this implies finding a propbank sense or a lemma. For that, we 
rely entirely on the AMR training set instead of using additional resources.

Given that the system runs two softmax operations, one to predict the action to take and the second one 
to predict the corresponding AMR node, and they both share LSTMs to make predictions, we include 
an additional layer with a \emph{tanh} nonlinearity after $\mathbf{s}_t$ for 
each softmax. 

\subsection{Word Representations}

We use character-based representations
 of words using bidirectional LSTMs \cite{ling:2015,ballesteros-dyer-smith:2015:EMNLP}. They
  learn  representations for words that are orthographically similar. Note that they are updated with 
  the updates to the model.
  \newcite{ballesteros-dyer-smith:2015:EMNLP} and  \newcite{lample2016} demonstrated 
   that it is possible to achieve high results in syntactic
  parsing and named entity recognition by just using character-based word representations 
  (not even POS tags, in fact, in some cases the results with just character-based representations
  outperform those that used explicit POS tags since they provide similar vectors for words with 
  similar/same morphosyntactic tag \cite{ballesteros-dyer-smith:2015:EMNLP});  in this paper we show 
  a similar result given that both syntactic parsing and named-entity recognition play a central role in AMR 
  parsing. 
  
 These are concatenated with pretrained word  embeddings. We use a variant of the skip n-gram model 
 provided by  \newcite{Ling:2015:naacl} with the LDC English Gigaword corpus (version 5). 
 These embeddings encode the syntactic behavior of the words (see 
 \cite{Ling:2015:naacl}).
 
 More formally, to represent each input token, we concatenate two vectors: a learned
 character-based representation ($\tilde{\mathbf{w}}_{\textrm{C}}$); and a 
fixed vector representation from a neural language model ($\tilde{\mathbf{w}}_{\textrm{LM}}$).
A linear map ($\mathbf{V}$) is applied to the resulting vector and passed through a component-wise ReLU,
\begin{align*}
\mathbf{x} = \max\left\{\mathbf{0}, \mathbf{V}[\tilde{\mathbf{w}}_{\textrm{C}}; \tilde{\mathbf{w}}_{\textrm{LM}}]  + \mathbf{b} \right\} .
\end{align*}
where $\mathbf{V}$ is a learned parameter matrix, $b$ is a bias term and ${\mathbf{w}}_{\textrm{C}}$
is the character-based learned representation for each word, $\tilde{\mathbf{w}}_{\textrm{LM}}$ is the 
pretrained word representation.

 \subsection{POS Tagging and Dependency Parsing}
 
 We may include preprocessed POS tags or dependency parses to incorporate more information into our model.  
For the POS tags we use the Stanford tagger \cite{Toutanova:2003:FPT:1073445.1073478} while we use the 
 \newcite{lstmacl15}'s Stack-LSTM parser trained on the English CoNLL 2009 
 dataset \cite{Hajic:2009:CST:1596409.1596411} to get the dependencies.
 

 \paragraph{POS tags:} The POS
  tags are preprocessed and a learned representation $\mathbf{tag}$ is concatenated with the word 
  representations. This is the same setting as \cite{lstmacl15}.

\paragraph{Dependency Trees:}  We use them in the same way as POS tags by concatenating a 
learned representation $\mathbf{dep}$ of the dependency label to the parent with the word 
representation. Additionally, we enrich the state representation $\mathbf{s}_t$, 
 presented in Section \ref{staterep}. If the two words at the top of the \textsc{stack} have  a 
 dependency between them, $\mathbf{s}_t$ is enriched 
 with a learned representation that indicates that and the 
direction; otherwise $\mathbf{s}_t$ remains unchanged.
 $\mathbf{s}_t$ is calculated as follows:
\begin{align*}
\mathbf{s}_t = \max \left\{\mathbf{0}, \mathbf{W}[\mathbf{st}_t; \mathbf{b}_t; \mathbf{a}_t; \mathbf{dep}_t] + \mathbf{d}\right\},
\end{align*}

\noindent where $\mathbf{dep}_t$ is the learned vector that represents that there is an 
arc between the two top words at the top of the stack.
 
\section{Experiments and Results}

We use the LDC2014T12 dataset\footnote{This dataset is a standard for comparison and has been used for 
evaluation in recent papers like \cite{WangXP15, 
goodman-vlachos-naradowsky:2016:P16-1,zhouemnlp}. We use the standard training/development/test
split: 10,312 sentences for training, 1,368 sentences
for development and 1,371 sentences held-out for
testing. } for our experiments. Table \ref{results} shows results, 
including comparison with prior work that are also evaluated on the same 
dataset.\footnote{The first entry for Damonte et al. is calculated using a pretrained LDC2015 
model, available at \url{https://github.com/mdtux89/amr-eager}, but evaluated on the LDC2014 dataset. 
This means that the score is not directly comparable with the rest. The second entry (0.64)
for Damonte et al. is calculated by training their parser with the LDC2014 training set which makes it directly 
comparable with the rest of the parsers.} 

\begin{table}[!ht]
\centering
\scalebox{0.73}{
\begin{small}
\begin{tabular}{l|c|c}
\textbf{Model} & \textbf{F}${_{\mathbf{1}}}$(Newswire) & \textbf{F}${_{\mathbf{1}}}$(ALL) \\
\hline
\newcite{jamr}* (POS, DEP)& 0.59 & 0.58 \\
\newcite{flanigan2016cmu}* (POS, DEP, NER)& -- & \bf0.66 \\
\newcite{stanfordamr}* (POS, DEP, NER) & 0.62 & -- \\
\newcite{DamonteCS16}$^8$(POS, DEP, NER, SRL) & -- & 0.61\\
\newcite{DamonteCS16}$^8$(POS, DEP, NER, SRL) &  -- & 0.64\\
\newcite{artzi-lee-zettlemoyer} (POS, CCG) & 0.66 & --\\
\newcite{goodman-vlachos-naradowsky:2016:P16-1}* (POS, DEP, NER)& 0.70 & --\\
\newcite{zhouemnlp}* (POS, DEP, NER, SRL) & \bf0.71 & \bf0.66 \\
\hline
\newcite{pust-EtAl:2015:EMNLP} (LM, NER)& -- & 0.61\\
\newcite{pust-EtAl:2015:EMNLP} (Wordnet, LM, NER)& -- &\bf0.66 \\
\hline
\newcite{wangxuepradhan}* (POS, DEP, NER)  & 0.63 & 0.59 \\
\newcite{WangXP15}* (POS, DEP, NER, SRL) &  0.70 &\bf0.66 \\
\hline
\hline
\textsc{Our parser (No pretrained-No chars)} & 0.64 & 0.59 \\
\textsc{Our parser (No pretrained-With chars)} & 0.66 & 0.61 \\
\textsc{Our parser (With pretrained-No chars)} & 0.66 & 0.62 \\
\hline
\textsc{Our parser} & 0.68 & 0.63 \\
\textsc{Our parser (POS)} & 0.68 & 0.63 \\
\textsc{Our parser (POS, DEP)} & 0.69 & 0.64 \\
\end{tabular}
\end{small}
}
\caption{AMR results on the LDC2014T12 dataset; Newsire section (left) and full (right). 
Rows labeled with \textsc{Our-Parser} show our results. POS indicates that the system uses preprocessed POS tags, 
DEP indicates that it uses preprocessed dependency trees, 
SRL indicates that it uses preprocessed semantic roles,  NER indicates that it uses preprocessed named entitites. LM indicates that it uses a LM trained on AMR data and WordNet 
indicates that it uses WordNet to predict the concepts. Systems marked with * are pipeline systems that require a
dependency parse as input. \textsc{(With pretrained-No chars)} shows the results of our parser 
without character-based representations. \textsc{(No pretrained-With chars)} shows results without 
pretrained word embeddings. \textsc{(No pretrained-No chars)} shows results without  character-based 
representations and without pretrained word embeddings. The rest of our results include both pretrained 
embeddings and character-based representations.}
\label{results}
\end{table}

Our model achieves 0.68 F1 in the newswire section of the test set just by using character-based 
representations of words and pretrained word embeddings. All prior work uses lemmatizers, 
POS taggers, dependency parsers, named entity recognizers and semantic role labelers that 
use additional training data while we achieve competitive scores without that. 
\newcite{pust-EtAl:2015:EMNLP} reports 0.66 F1 in the full test by using WordNet for concept identification,
 but their performance drops to 0.61 without WordNet. It is worth noting that we achieved 
 0.64 in the same test set without WordNet.
\newcite{wangxuepradhan,WangXP15} without SRL (via Propbank) achieves only 0.63 in 
the newswire test set while we achieved 0.69 without SRL (and 0.68 without dependency 
trees).

In order to see whether pretrained word embeddings and character-based embeddings are useful we
carried out an ablation 
study by showing the results of our parser with and without 
character-based representations (replaced by standard lookup table learned embeddings) 
and with and without pretrained word embeddings. 
By looking at the results of the parser without character-based embeddings but with pretrained word 
embeddings we observe that the character-based representation of words are useful since they help to 
achieve 2 points better in the Newswire dataset and 1 point more in the full 
test set. The parser with character-based embeddings but without 
pretrained word embeddings, the parser has more difficulty to learn and only 
achieves 0.61 in the full test set. Finally, the model that does not use neither 
character-based embeddings nor pretrained word embeddings is the worst achieving only 0.59 in the 
full test set, note that this model has no explicity way of getting any syntactic information through the 
word embeddings nor a smart way to handle out of vocabulary words. 

All the systems marked with * require that the input is a dependency tree, which means that they solve a 
transduction task between a dependency tree and an AMR graph.  Even though our parser starts from
plain text sentences when we incorporate more information into our model, we achieve further 
improvements. POS tags provide small improvements (0.6801 without POS tags vs 0.6822 
for the model that runs with POS tags). Dependency trees help a bit more achieving 0.6920.





\section{Conclusions and Future Work}

We present a new transition-based algorithm for AMR parsing and we implement
it using Stack-LSTMS and a greedy decoder. We present competitive results, without any additional 
resources and  external tools. Just by looking at the words, we achieve 0.68 F1 (and 0.69 by 
preprocessing dependency trees) in the standard dataset used for evaluation.


\section*{Acknowledgments}
 We thank Marco Damonte, Shay Cohen and Giorgio Satta for running and evaluating their parser 
 in the LDC2014T12 dataset. We also thank Chuan Wang for useful discussions.

\bibliography{acl2017,main}
\bibliographystyle{emnlp_natbib}

\end{document}